\documentclass[conference]{IEEEtran}
\IEEEoverridecommandlockouts
\usepackage{cite}
\usepackage{amsmath,amssymb,amsfonts}
\usepackage{algorithmic}
\usepackage{graphicx}
\usepackage{textcomp}
\usepackage{xcolor}
\usepackage{comment}
\usepackage{multirow}
\def\BibTeX{{\rm B\kern-.05em{\sc i\kern-.025em b}\kern-.08em
    T\kern-.1667em\lower.7ex\hbox{E}\kern-.125emX}}
\begin{document}

\title{GatedxLSTM: A Multimodal Affective Computing Approach for Emotion Recognition in Conversations\\}

    
\author{Yupei Li, Qiyang Sun, Sunil Munthumoduku Krishna Murthy, Emran Alturki, and Björn W. Schuller \IEEEmembership{Fellow, IEEE}

\thanks{Yupei Li, Qiyang Sun and Emran Alturki are with GLAM, Department of Computing, Imperial College London, UK (e-mail:yupei.li22@imperial.ac.uk;q.sun23@imperial.ac.uk;  e.alturki24@imperial.ac.uk).}
\thanks{Sunil Munthumoduku Krishna Murthy is with CHI -- Chair of Health Informatics, MRI, Technical University of Munich, Germany (e-mail: sunil.munthumoduku@tum.de).}
\thanks{Björn W.\ Schuller is with GLAM, Department of Computing, Imperial College London, UK; CHI -- Chair of Health Informatics, Technical University of Munich, Germany; relAI -- the Konrad Zuse School of Excellence in Reliable AI, Munich, Germany; MDSI -- Munich Data Science Institute, Munich, Germany; and MCML -- Munich Center for Machine Learning, Munich, Germany (e-mail: schuller@tum.de).}
\thanks{Yupei Li and Qiyang Sun contributed equally to this work.}
}

\maketitle

\begin{abstract}
Affective Computing (AC) is essential for advancing Artificial General Intelligence (AGI), with emotion recognition serving as a key component. However, human emotions are inherently dynamic, influenced not only by an individual’s expressions but also by interactions with others, and single-modality approaches often fail to capture their full dynamics. Multimodal Emotion Recognition (MER) leverages multiple signals but traditionally relies on utterance-level analysis, overlooking the dynamic nature of emotions in conversations. Emotion Recognition in Conversation (ERC) addresses this limitation, yet existing methods struggle to align multimodal features and explain why emotions evolve within dialogues. To bridge this gap, we propose GatedxLSTM, a novel speech-text multimodal ERC model that explicitly considers voice and transcripts of both the speaker and their conversational partner(s) to identify the most influential sentences driving emotional shifts. By integrating Contrastive Language-Audio Pretraining (CLAP) for improved cross-modal alignment and employing a gating mechanism to emphasise emotionally impactful utterances, GatedxLSTM enhances both interpretability and performance. Additionally, the Dialogical Emotion Decoder (DED) refines emotion predictions by modelling contextual dependencies. Experiments on the IEMOCAP dataset demonstrate that GatedxLSTM achieves state-of-the-art (SOTA) performance among open-source methods in four-class emotion classification. These results validate its effectiveness for ERC applications and provide an interpretability analysis from a psychological perspective.
\end{abstract}


\section{Introduction}
Artificial General Intelligence (AGI) represents a key future direction in AI development, with Affective Computing (AC) playing a crucial role in enhancing AGI's ability to interact effectively with humans. Sun et al.\ \cite{sun2024friendlyaicomprehensivereview} advocate this perspective, emphasising that emotional recognition is a fundamental and essential technique within AC. Emotional recognition has been extensively explored in the literature. For instance, Venkataramanan et al.\ \cite{venkataramanan2019emotionrecognitionspeech} examine various audio features to predict emotions, while text-based emotion recognition has also been developed \cite{al2024challenges}. However, relying on a single modality is insufficient for capturing emotions, as they are expressed through audio, text, and visual cues from different perspectives \cite{zhang2024deep}.
Consequently, Multimodal Emotion Recognition (MER) has gained attention for its ability to incorporate diverse emotional expressions, for example audio and text, in order to mitigate misinterpretation.

MER encompasses various model types, which differ based on the granularity of the corpus. Word-level MER models, for instance, focus on short phrases. Local attention mechanisms have been applied to frame-level acoustic features in Recurrent Neural Networks (RNNs) \cite{7952552}, and word clustering models have also been utilised \cite{yuan2016use}. These approaches often enhance recognition speed by prioritising key emotional words, which is particularly advantageous for real-time MER systems requiring immediate responses. However, such models may sacrifice a comprehensive understanding of contextual or situational factors. Utterance-level models address this limitation by leveraging entire sentences for emotion prediction \cite{zhang2024survey}. Nevertheless, human emotions are inherently dynamic, exhibiting momentary and situational variations \cite{Izard2009}. Relying solely on utterance-level representations may result in the loss of critical contextual information, thereby failing to capture the evolving nature of emotions. To mitigate this issue, Emotion Recognition in Conversation (ERC) has been explored through AC models \cite{hu2024recent}, which incorporate broader contextual background information alongside the current utterance, thereby enhancing the accuracy of emotion recognition.

Despite advancements in MER within ERC, several challenges remain. First, multimodal fusion methodologies have not been universally effective. Extensive research has explored early, late, and model fusion strategies; however, the integration of features across modalities remains suboptimal, particularly due to alignment issues \cite{zhu2024review}. Although cross-attention mechanisms \cite{khan2025memocmt} have been employed for alignment, they primarily adapt one modality to another rather than achieving comprehensive fusion. Additionally, the incorporation of contextual information in conversations remains limited. Many approaches simply concatenate features from previous utterances without explicitly determining their contributions to emotion 
\bstctlcite{IEEEexample:BSTcontrol}
recognition \cite{messaoudi2024modeling}. Furthermore, ERC is inherently a sequential classification problem, typically addressed using time-series models. While Long Short-Term Memory (LSTM)
and RNN architectures \cite{messaoudi2024speech} are 
commonly employed, they struggle to retain information over long sequences, as their processing tends to prioritise recent states. Transformer-based architectures \cite{hazmoune2024using} mitigate this limitation by leveraging self-attention mechanisms, yet, they do not inherently emphasise the sequential order of data to the same extent as traditional time-series models.

To address these challenges, Contrastive Language-Audio Pretraining (CLAP) has been introduced for fusing text and audio modalities by utilising a shared embedding space, which we adopt in this study \cite{laionclap2023}. Additionally, xLSTM has been proposed as a hybrid approach that integrates the advantages of both LSTMs and Transformers; however, it has not yet been applied to this domain \cite{beck2024xlstm}. Although large language models (LLMs) have been explored for ERC \cite{dutta2025llmsupervisedpretrainingmultimodal}, their computational cost remains prohibitively high. Consequently, this study focuses on deep learning models that offer a more practical balance between performance and efficiency. Furthermore, to address the issue of contextual dependency, we propose a novel gate-based architecture that explicitly determines the contributions of different utterances within a conversation. This mechanism enhances interpretability and provides structured guidance for the model to refine its focus. Additionally, we find the Dialogical Emotion Decoder (DED) \cite{9053561} to be a useful approach for addressing this challenge.

Our contributions are as follows:
\begin{itemize}
   \item Utilise an xLSTM-based model in the ERC task firstly to the best of our knowledge. Propose a model-level fusion GatedxLSTM for ERC within AC, demonstrating state-of-the-art (SOTA) performance among open-source models for four-class emotion classification on the IEMOCAP dataset.
    \item Enhance the interpretability of the ERC task by analysing the importance of the current speaker and the interlocutor in emotion prediction.
    \item Outline future research directions for ERC, highlighting potential advancements and challenges in the field.
\end{itemize}

\section{Related Work}
\begin{figure*}
    \centering
    \includegraphics[width=\linewidth]{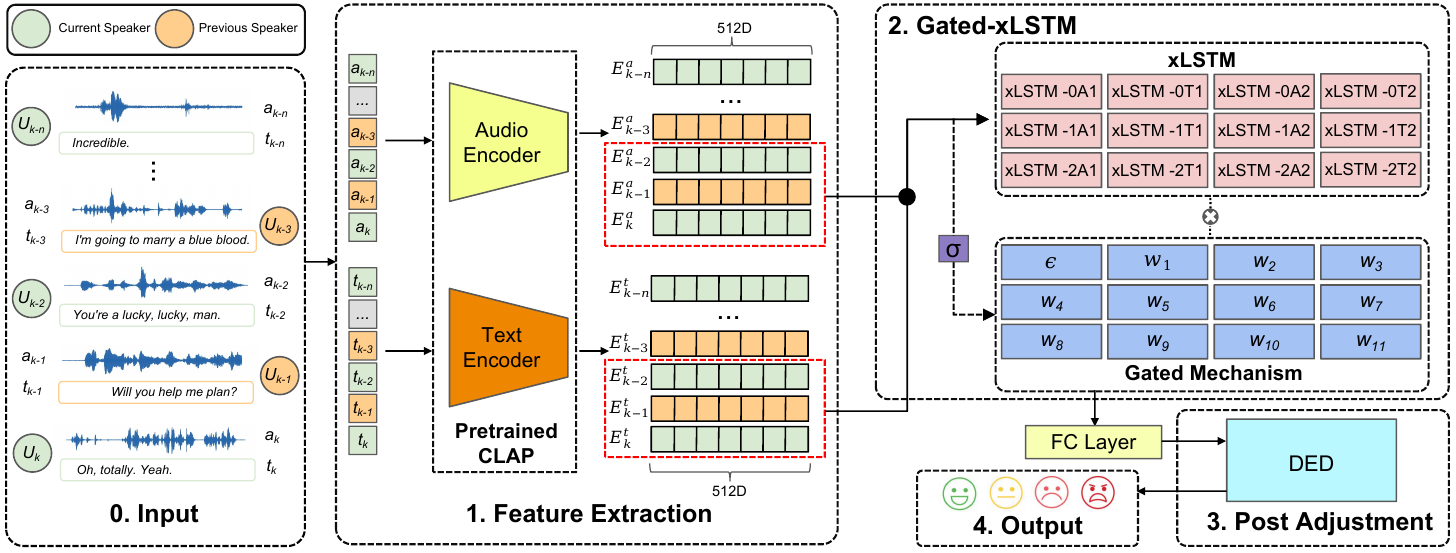}
    \caption{The pipeline of GatedxLSTM.  First, each audio sample and its corresponding transcription are processed using the CLAP model, which maps them into a shared embedding space and aligns their embeddings. Next, for a given utterance at time $T$, we identify its preceding utterance spoken by the interlocutor. Both the audio and text representations are then passed through four distinct xLSTM blocks to extract features. To incorporate contextual information, we retrieve relevant features from several preceding utterances (e.\,g., at $T-1$). These extracted features are then processed through a gating mechanism to determine their contribution to the final emotion recognition task, assigning the audio of the current utterance a weight of 1. After applying the gating mechanism, the features are passed through a fully connected layer. Finally,
    the prediction is refined using DED in the last stage of the pipeline.}
    \label{fig:pipeline}
    \vspace{-0.3cm}
\end{figure*}
\subsection{Emotion Recognition in Conversation (ERC)}

Research shows that emotions can be conveyed through multiple channels, including vocal prosody, linguistic content, visual expressions, and physiological signals \cite{keltner2019emotional,de2015perception}. MER integrates information from different modalities to improve recognition accuracy. Compared to single-modal approaches, multimodal interaction captures emotional characteristics more comprehensively and enhances model robustness \cite{zhu2024review}. Existing studies adopt different modality combinations to optimise feature representation. Wang et al.\ \cite{wang2022multi} developed the MED4 multimodal database integrating electroencephalogram (EEG), speech, video, and photoplethysmography. Their feature-level and decision-level fusion framework demonstrated enhanced recognition accuracy in noisy environments through EEG-speech synergy. This revealed complementary interactions between physiological and behavioural signals. Middya et al.\ \cite{MIDDYA2022108580} employed a deep convolutional network to extract MFCC features from audio and spatiotemporal features from video. They applied model-level fusion to develop a lightweight multimodal model, achieving 86\% and 99\% accuracy on the RAVDESS and SAVEE datasets, respectively. Yin et al.\ \cite{yin2024token} explored the combination of speech, video, and text, proposing the Token-Disentangling Mutual Transformer (TMT) for cross-modal feature interaction. Their approach first disentangles emotion-related features across modalities and then applies bidirectional query learning within a Transformer framework for feature fusion. Experiments on the CMU-MOSI, CMU-MOSEI, and CH-SIMS datasets achieved SOTA performance in single-utterance emotion recognition. While these methods have contributed to MER, most research still focuses on utterance-level emotion classification. They do not consider the evolution of emotional states in conversations \cite{lian2023survey}.

ERC has been proposed to capture the evolution of emotions in dialogue \cite{poria2019emotion}. It emphasises contextual modelling to track how emotions change throughout an interaction. Currently, various studies have explored multimodal approaches to model emotions in conversation. Meng et al.\ \cite{meng2024revisiting} proposed the Graph-Spectrum-based Multimodal Consistency and Complementary Learning Framework (GS-MCC), integrating speech, text, and video for ERC. This method constructs a multimodal interaction graph and employs a Fourier Graph Neural Network to extract high- and low-frequency components over long-range dependencies. It enhances consistency and complementarity across modalities using contrastive learning, improving classification accuracy. Xie et al.\ \cite{xie2021robust} introduced the Crossmodality Transformer, a fusion model for multimodal emotion recognition. It incorporates the EmbraceNet structure and applies the Transformer mechanism to integrate speech, video, and text features. Experiments on the MELD dataset demonstrate its effectiveness, surpassing unimodal models in performance. The model captures contextual dependencies in conversations, improving emotion recognition accuracy. Experiments on the MELD dataset, which contains multi-turn dialogues, demonstrate its effectiveness, outperforming unimodal models. Wang et al.\ \cite{9533718} proposed the Contextual Attention Network (CAN) for multimodal ERC, combining speech, text, and video information. This model is based on a Gated recurrent units (GRU) network to capture intra- and inter-speaker emotional dynamics. It integrates BERT for text embeddings, a Convolutional Neural Network (CNN) for speech processing, and 3D-CNN for video analysis. Further, an attention mechanism is designed to refine cross-modal weight adjustments. CAN achieved 64.6\% accuracy on the IEMOCAP dataset, outperforming multiple baselines.

While these methods demonstrate the feasibility of contextual modelling in ERC, multimodal ERC still faces challenges. One issue is modalities'
fusion, as the effective integration of multiple modalities while minimising redundancy and conflict remains unresolved \cite{gandhi2023multimodal}.

\subsection{Contrastive Language-Audio Pretraining (CLAP)}

MER relies on feature extraction from different modalities. In speech and text-based scenarios, traditional methods typically model these modalities independently. The speech modality often uses Low-Level Descriptors (LLD) or pretrained model-based high-dimensional features. The text modality relies on word embeddings or pretrained language models for semantic representation. Since these features are extracted in separate spaces, semantic misalignment between speech and text representations occurs; consequently, it affects the effectiveness of multimodal fusion \cite{zhang2024deep}.

CLAP \cite{elizalde2023clap} is designed to  align the representations of audio and language. It applies contrastive learning to project speech and text into a shared embedding space, improving cross-modal alignment. CLAP follows the principles of Contrastive Language-Image Pretraining (CLIP) \cite{radford2021learning} and aims to maximise the similarity between matching speech-text pairs while minimising the similarity between mismatched samples. It uses a dual-encoder architecture, employing a speech encoder and a text encoder to process inputs separately before computing their similarity in a shared feature space.  

Some studies explore the use of CLAP in emotion-related tasks. ParaCLAP \cite{jing2024paraclap} focuses on paralinguistic analysis, including speech emotion recognition (SER). It enhances feature alignment through speech-text contrastive learning and constructs speech-text pairs using emotion labels and expert features. The model learns cross-modal representations in a shared space, focusing primarily on optimising data construction rather than modifying CLAP’s pretraining strategy. GEmo-CLAP \cite{pan2024gemo} applies CLAP to SER, incorporating contrastive learning with Kullback-Leibler (KL) divergence loss to refine the speech-text similarity matrix. It also introduces gender information and employs multi-task learning and soft label training to improve SER performance. 

Unlike these methods, we utilise CLAP as a feature extraction and alignment tool. This ensures alignment of features from these text and audio
modalities into approximately the same feature space. The extracted features serve as inputs to our proposed Gated-xLSTM model.

\subsection{xLSTM for Sequential Modelling}

The traditional LSTM architecture faces limitations in capturing long-range dependencies, enforcing strict sequential computation, and handling constrained memory storage \cite{waqas2024critical}. To address these issues, Extended LSTM (xLSTM) \cite{beck2024xlstm} enhances model expressiveness and computational efficiency through exponential gating and an improved memory structure. xLSTM introduces two key memory mechanisms: Scalar Memory LSTM (sLSTM) and Matrix Memory LSTM (mLSTM). sLSTM utilises multi-head memory units, improving hierarchical information modelling and enabling better feature disentanglement. mLSTM utilises a matrix memory structure to store key information and applies a covariance update rule to enhance long-term memory retention. It also supports parallel computation, reducing computational bottlenecks. These improvements make xLSTM more effective in long-sequence modelling, cross-time-step information storage, and computational parallelism. Therefore, xLSTM is particularly suitable for multi-turn dialogues, reinforcement learning, and temporal reasoning tasks.

Several studies demonstrate the  advantages of xLSTM in long-term dependency modelling. Schmidinger et al.\ \cite{schmidinger2024bio} proposed Bio-xLSTM for genomics, protein structure analysis, and small molecule modelling. Their findings show that xLSTM’s exponential gating mechanism effectively captures long-range dependencies, improving DNA and RNA sequence modelling accuracy. Zendehbad et al.\ \cite{zendehbad2025traxvbf} introduced the TraxVBF framework in electromyography (EMG) processing, combining xLSTM and Transformers for neurological rehabilitation and motion prediction. The study shows that xLSTM outperforms conventional LSTM and GRU
in motion trajectory forecasting and assistive technology applications. Huang et al.\ \cite{huang2024xlstm} applied xLSTM-FER to facial expression recognition (FER). The model processes facial feature patches using xLSTM, establishing temporal dependencies across time steps. It achieves superior performance over CNNs and Vision Transformers (ViT)
on the CK+, RAF-DB, and FERplus datasets. Derington et al.\ \cite{derington2025testing} evaluated xLSTM and other acoustic models in SER. Their study focuses on model robustness and fairness across different emotional dimensions.

Although these studies highlight xLSTM’s strength in long-term dependency modelling, its potential remains underexplored in AC tasks, particularly in multimodal ERC.

\subsection{Dialogical Emotion Decoding (DED)}
DED \cite{yeh2020dialogical} is an approximate inference algorithm that refines dialogue-based emotion prediction through sequential decoding and contextual rescoring. Unlike models that classify utterances independently, DED incorporates historical emotional patterns to improve prediction accuracy. It follows three key principles: 1) frequently occurring emotions in a dialogue are more likely to persist, 2) posterior probabilities are dynamically updated rather than relying on rigid labels, and 3) the emotional states of different speakers are interdependent and should be modelled jointly. Given predictions from a pre-trained emotion recognition model, DED refines results in the following steps. First, emotion shifts are modelled using a Bernoulli distribution. Then, emotion states are clustered with a distance-dependent Chinese Restaurant Process (ddCRP) \cite{blei2011distance}. Finally, Beam Search \cite{steinbiss1994improvements} is applied to ensure a contextually coherent prediction sequence. This post-processing approach has been widely validated in recent ERC research \cite{zhang2024multimodal, atkar2023speech, chen2022emotion}.

\section{Methodology}

Based on the motivation and the identified gaps, we propose our pipeline model, as illustrated in Figure \ref{fig:pipeline}. Each stage of the model is explained in detail in the subsequent sections. 

\subsection{Feature extraction}
Effective alignment across different modalities is crucial, as misalignment can lead to suboptimal feature extraction. To explore this, we conducted preliminary experiments on the IEMOCAP dataset (discussed in Section \ref{sec:dataset}), evaluating uni-modal performance and simple concatenation using an xLSTM model, as presented in Table \ref{tab:preliminary}. The results indicate that 
simple concatenation outperforms both uni-modal approaches, confirming that integrating multiple modalities helps compensate for limitations in individual modalities. However, this approach still falls short of SOTA methods, highlighting that simple concatenation is insufficient for effectively fusing modalities while preserving alignment features.

\begin{table}[htbp]
    \centering
    \caption{Performance of LSTM and xLSTM models with different modality and context features. Multimodal fusion is implemented through simple concatenation, while context features are derived by interacting the current utterance with the features of the preceding and subsequent two sentences. Both methods of concatenation and an Interaction-aware Attention Network (IAAN) \cite{8683293} have been utilised as context features extraction.}
    \begin{tabular}{c c c c}
    \hline
    Model & Modality& Context & Accuracy (\%) \\
    \hline
       LSTM  & Audio + Text & No & 42.3 \\
       LSTM  & Audio + Text & Yes (concatenation) & 45.7  \\
       xLSTM & Audio & No & 56.1   \\
       xLSTM & Text & No & 63.7   \\
       xLSTM & Audio + Text & No & 66.2   \\
       xLSTM & Audio & Yes (IAAN) & 65.3  \\
       xLSTM & Text & Yes (IAAN) & 38.1  \\
       xLSTM & Audio + Text & Yes (IAAN) & 67.3   \\
       \hline
    \end{tabular}

    \label{tab:preliminary}
\end{table}
Instead, CLAP employs a pretraining strategy to align text and audio features, mapping them into a shared embedding space. Unlike cross-attention mechanisms, which allow features to influence each other implicitly, CLAP explicitly matches text and audio representations. This alignment has been shown to be beneficial for downstream classification tasks. Therefore, we utilise a CLAP pre-trained model to acquire the aligned features, shown in Equation \ref{eq:clap}, trained in contrastive loss function as discussed in \cite{laionclap2023}.
\begin{equation}
\label{eq:clap}
\begin{aligned}
    E^a_j &= \text{MLP}_{\text{audio}}\left(g_{\text{audio}}(A_j)\right) \\
    E^t_j &= \text{MLP}_{\text{text}}\left(g_{\text{text}}(T_j)\right) \\
\end{aligned}
\end{equation}
The $g_{\text{audio}}$ and $g_{\text{text}}$ are the audio and text encoder, along with the MLP, used as a projector to obtain the fixed shared embedding space representation. $A_j$ and $T_j$ are the raw audio and text from the dataset, and the corresponding $E^a_j$ and $E^t_j$ are the aligned embeddings.
\subsection{Gated-xLSTM}
\subsubsection{Base xLSTM}
As previously mentioned, the ERC task is a sequential modelling problem for which LSTM has demonstrated strong performance. However, attention mechanisms have also been introduced to enhance the model's ability to capture global contextual information. The xLSTM model effectively combines the advantages of both approaches while incorporating a memory matrix to improve memory efficiency compared to traditional Transformers. Therefore, we selected xLSTM as the base model. As shown in Table \ref{tab:preliminary}, xLSTM outperforms LSTM, further validating our choice.
\subsubsection{Gate mechanism}
While we have aligned feature extraction methods and sequence models for this task, we still lack a clear understanding of why we obtained final emotion classification results. Intuitively, as suggested by previous literature, preceding sentences are expected to influence the emotional state at the current stage. However, it remains unclear which modality from which contextual utterance exerts the dominant influence. Furthermore, we have not examined whether the interlocutor or the speaker themselves plays a greater role in shaping the current emotional state. Psychologically, evidence suggests that emotions influence both the self and others \cite{pekaar2020dynamic}, yet determining which has a more significant impact remains a complex challenge. To better capture the importance of different features, we design a gating mechanism that facilitates the selection and refinement of the aligned features we extract.
We first pass the aligned features to xLTSM shown in Equation \ref{eq:xlstm}, where $x_j$ represents audio or text features we extract. 
\begin{equation}
\label{eq:xlstm}
    x_j = \text{xLSTM}(E_j), \quad\text{where}\quad E_j \in \{E^a, E^v\}
\end{equation}

Notably, xLSTM comprises sLSTM and mLSTM blocks as its key components. We posit that the integration of LSTM and the attention mechanism in mLSTM addresses the challenges associated with each individual component discussed previously. This integration, in addition to the standard LSTM gating mechanism, is illustrated in Equation \ref{eq:mlstm} \cite{beck2024xlstm},
\begin{equation}
\label{eq:mlstm}
\begin{aligned}
    C_t &= f_t C_{t-1} + i_t v_t k_t^\top    \\
    n_t &= f_t n_{t-1} + i_t k_t   \\
    h_t &= o_t \odot \tilde{h}_t, \quad \tilde{h}_t = \frac{C_t q_t}{\max\left(n_t^\top q_t, 1\right)} \\
    q_t &= W_qE_{jt} + b_q, \quad k_t = \frac{1}{\sqrt{d}}W_kE_{jt} + b_k, \\
    \quad v_t &= W_vE_{jt} + b_v,\\
\end{aligned}
\end{equation}
where $C_t$ is the current cell state, $f_t$ is the forget gate, and $q, k, v$ follow the same definitions as in the attention mechanism. Additionally, $i_t$ is the input gate, $o_t$ is the output gate, $h_t$ is the hidden state, $n_t$ is the normalised state, and $E_i$ is the embedding obtained from CLAP. Notably, the integration of the attention mechanism into the cell and hidden state updates is a novel approach, enabling sequence modelling to capture the entire context more effectively.

After we get the output of xLSTM, inspired by the forget gate in LSTM, we hypothesise that neural networks have the capability to determine which features are important and should be retained for a longer duration when explicitly guided by emphasis and structured queries regarding their retention. Therefore, we adopt a similar strategy, enabling the model to learn to prioritise features based on the current context, as shown in Equation \ref{eq:forget}.
\begin{equation}
\label{eq:forget}
\begin{aligned}
    w_j &= \text{Sigmoid}\left(\text{Linear}(E_j)\right), \\
\end{aligned}
\end{equation}

where $w_j$ is the forget weight, ranging from 0 to 1, as a scalar value, obtained through a sigmoid activation function. 

These forget weights are then applied to multiple xLSTM layers to obtain sequence feature representations. The multiplication follows an element-wise extension, meaning the weights are applied across all dimensions of the feature modality for a given utterance, shown in Equation \ref{eq:forget2}.
\begin{equation}
\label{eq:forget2}
\begin{aligned}
z_{j} &= w_j \cdot x_j \\
y_j &= \text{Linear}(Z), \\
\end{aligned}
\end{equation}
where $x_j$ is each value in the hidden dimension, and we build $Z$ from concatenation of $z_j$. Also, $y_j$ is the predicted emotion classification. We do not consider the sequence length here, as we use the final token to represent the features of the sentence. Notably, we deliberately cluster the current utterance with the utterance spoken by the previous interlocutor at time $k$ to ensure they are processed together. It is possible that at time $k-1$, the utterance still belongs to the current speaker; however, we continue iterating through previous time frames to identify the utterance of the interlocutor. This approach is intended to compel the model to learn the importance of the four features: the audio and text, from both the speaker and the interlocutor. Additionally, we standardise the current utterance's audio by representing it as a unit vector.

Moreover, we take contextual information into account, although we focus only on a limited number of previous utterances. While subsequent sentences may provide additional valuable insights, such as maintaining emotional consistency, it is impractical to anticipate future utterances in real-time applications. The features of the previous utterances, each encompassing four distinct features per time frame, are passed together through the gate mechanism to assess their relative importance.

Subsequently, we perform another fusion of the features through simple concatenation via a fully connected layer, even though multimodal features were already fused in CLAP. This step aims to project the features $Z$, which have the shape (batch size $\times$ number of sentences $\times$ hidden dimension), into the appropriate classification space. Additionally, it enables model-level fusion of the xLSTM output features, providing the model with an additional opportunity to learn the interdependencies among the multimodal features.

\subsection{Post adjustment}

DED \cite{yeh2020dialogical} is employed for the post adjustment stage. Given a dialogue \( U = \{ u_1, u_2, ..., u_k \} \), the goal is to decode the
emotion state \( y_k \) of each utterance \( u_k \) based on the predicted emotion states \( Y_{1:k-1} \). The joint probability distribution is formulated as:
\begin{equation}
\label{eq:joint_prob}
\begin{aligned}
p(Y, S) = p(y_1 | x_1) \prod_{k=2}^{K} p(y_k | x_k) p(y_k, s_k | Y_{1:k-1}, S_{1:s-1}),
\end{aligned}
\end{equation}

where \( x_k \) is the feature vector of \( u_k \), \( y_k \in \{\text{anger}, \text{happiness}, \text{neutrality}, \text{sadness} \} \) represents the predicted emotion label, and \( S = \{s_1, ..., s_K\} \) is a sequence of binary variables where each \( s_k \) indicates whether an emotion shift occurs at utterance \( u_k \).


Emotion shifts follow a Bernoulli distribution with probability \( p_0 \), which is estimated from training data based on observed emotion transitions.



Emotion assignment is performed using a ddCRP. In this process, the probability of assigning an emotion label \(y_k\)  depends on whether the emotion has appeared in the previous utterances. If emotion \(l\) has already been observed in earlier utterances, its assignment probability is proportional to the size of the corresponding emotion block, denoted as \( N_{l,k-1} \). Otherwise, a new emotion state is introduced with a probability controlled by the hyperparameter  \( \alpha \), which defaults to 1.

The optimal emotion sequence is determined by maximising the posterior probability:
\begin{equation}
\label{eq:decoding}
Y^* = \arg\max_Y \log p(X, Y),
\end{equation}
where \( p(X, Y) \) incorporates probabilities from the emotion classification module, emotion shift modelling, and emotion assignment. Finally, Beam Search is selected for decoding, narrowing down the search to the most promising sequences. This approach refines emotion predictions and aligns them with the natural flow of conversation.

\section{Experiment}
\subsection{IEMOCAP Dataset}
\label{sec:dataset}

IEMOCAP \cite{busso2008iemocap} is a multimodal database designed for emotion recognition and analysis. It comprises speech, video, and emotion annotations provided by three independent annotators. The dataset includes five sessions, each featuring a dyadic interaction between a male and a female actor. These sessions contain both scripted and improvised dialogues. In this study, we focus on four emotion categories
as commonly done on this database: 
happiness, neutrality, anger, and sadness. The dataset we filtered consists of 151 dialogues and 4,490 utterances, distributed as follows: happiness (13.25\%), neutrality (38.04\%), anger (24.57\%), and sadness (24.14\%).  

In this study, we utilise both textual and acoustic modalities from the IEMOCAP dataset, while the features are provided in the original dataset. Given that our experimental design incorporates contextual information, we employ a dialogue-level data split to prevent data leakage. Specifically, we ensure that sentences from the same dialogue appear within the same training, validation or test subset. The dataset is partitioned into training, validation, and test sets in an 8:1:1 ratio, with a random seed of 42 to split data for reproducibility. Additionally, we provide the dataset in the supplementary materials to facilitate replication.
The final results represent the average of five independent experiments with different seeds.
\subsection{Implementation and results}
Our experiments were conducted on a \emph{V100-PCIE-32GB} GPU. For CLAP, we utilised the default pre-trained model from the \emph{laion\_clap} library, keeping it frozen throughout the entire training pipeline. This decision was made as the model has been trained on a large-scale dataset, and we prioritised allocating our limited computational resources to further training the xLSTM component. The parameters for xLSTM are detailed in Table \ref{tab:parameter}. While xLSTM requires a sequential input, the complete representation from CLAP is not inherently sequential. To approximate the temporal structure of the audio clip, we segment the representation into 16 time steps, ensuring an even distribution across the sequence. Moreover, to incorporate contextual information from preceding utterances, we selected the two most recent utterance blocks, thus providing us 12 xLTSM blocks.

\begin{table}[htbp]
    \caption{Parameters and their value in our experiments}
    \centering
    \begin{tabular}{c|c}
    \hline
       Parameter  & Value \\
       \hline
      CLAP output dimension & 512 \\
      Batch size   & 32 \\
      xLSTM layers & 8 \\
      Maximum sequence length &  256 \\
      sLSTM \& mLSTM kernel size & 4 \\
      sLSTM \& mLSTM attention head & 4 \\
      Projection block size for QKV & 4 \\
      Activation & GELU \\
      Projection factor in feedforward & 1.3 \\
      Optimiser & Adam \\
      Learning rate & 0.001 \\
\hline      
    \end{tabular}

    \label{tab:parameter}
\end{table}

We use seed values of 42, 43, 45, 46, and 50 to obtain the final weighted accuracy and weighted F1-score, which are then compared against existing open-source SOTA models. These two metrics are selected due to the class imbalance inherent in the dataset. Weighted accuracy ensures that each class contributes proportionally to the overall performance evaluation, to get us equal evaluation of each class. Similarly, the weighted F1-score accounts for both precision and recall across imbalanced classes, providing a more comprehensive assessment of model performance.
\begin{table*}[htbp]
    \caption{Comparison of weighted Accuracy and F1 score with other models on IEMOCAP.}
    \centering
    \resizebox{0.9\textwidth}{!}{
    \begin{tabular}{c c c c c c}
    \hline
    Model & Modality & Classes & Open Source & W-Acc (\%) & W-F1 (\%) \\
    \hline
    FATRER \cite{mao2023fatrer}     & Text  & 6  & Yes & 69.69 & 69.35 \\
    BiosERC \cite{xue2024bioserc}   & Text  & 9  & Yes & -     & 71.19 \\
    ESA-CRF \cite{chen2022emotion}  & Audio & 4* & No  & 73.17 & - \\
    BoAW-RNN-ATT \cite{chamishka2022voice}  & Audio & 6 & No & 60.87 & 61.01 \\
    AF-CAN \cite{wang2021contextual}& Audio+Text+Video & 6 & No & 64.63 & 63.71 \\
    MPT-HCL \cite{zou2023multimodal}& Audio+Text+Video & 6 & No & 72.83 & 72.51 \\
    Attention-FCN \cite{zhang2018attention} & Audio+Text+Video & 4 & No & 70.40 & - \\
    ResNet+BERT \cite{padi2022multimodal}  & Audio+Text & 4* & No & 75.01 & - \\
    MER-HAN \cite{zhang2023multimodal}     & Audio+Text & 4* & No & 73.33 & 73.66 \\
    DPN-KL \cite{wu2023distribution}       & 
    Audio+Text & 4* & No & \emph{77.83} & - \\
    \textbf{GatedxLSTM (Ours)}             & Audio + Text & 4  & Yes & \textbf{76.34 $\pm$ 1.31} & \textbf{75.97 $\pm$ 1.38} \\
    \hline
    \end{tabular}
    }
    \label{tab:sota}
    
    \vspace{0.25 em}
    \footnotesize{* \textbf{Excitement} and \textbf{Happiness} categories are merged into a single \textbf{happiness} category in their study.}
    \vspace{-0.3cm}

\end{table*}

\section{Discussion}
Our proposed GatedxLSTM model, which leverages both audio and text modalities, achieves superior performance with a weighted
accuracy of 76.34$\pm$1.31\% and a weighted F1-score of 75.97$\pm$1.38\%.  These results highlight the effectiveness of our model in multimodal learning. Furthermore, our approach achieves SOTA performance among open-source models and approaches SOTA performance in general, with the exception of \cite{wu2023distribution}.

To further assess the effectiveness of our model, we evaluate its performance across individual emotion subclasses, as presented in Table \ref{tab:each}. The reported results reflect the average scores from five independent experiments. Our model demonstrates consistently high performance in classifying Neutral, Sad, and Angry emotions, indicating its reliability in these categories. However, performance decreases in the Happy category, suggesting inherent challenges in recognising happiness. We hypothesise that happiness may be more difficult to detect because individuals often feign happiness for social politeness rather than expressing genuine emotion \cite{crivelli2018facial}. Additionally, research suggests that happiness diminishes over time in response to sustained positive stimuli, a phenomenon known as the Hedonic Treadmill \cite{byrnes2005hedonic}. Consequently, incorporating additional feature extraction techniques or data augmentation strategies may be necessary to enhance the model’s ability to recognise Happy emotions accurately.

Moreover, certain studies, such as \cite{wu2023distribution}, report higher performance by redefining the happiness category to include \emph{excitement}. From a psychological perspective, however, this classification is questionable. While both happiness and excitement are positive emotions, excitement exhibits a more explicit expression and is often elicited by novel or stimulating experiences. According to the circumplex model of affect \cite{russell1980circumplex}, excitement is characterised by high arousal and lower in valence, whereas happiness is associated with lower
arousal. Therefore, although merging these two categories may improve classification accuracy, we argue that treating happiness and excitement as a single class does not align with theoretical expectations.

\begin{table}[h] 

\caption{Performance of the GatedxLSTM model per class}
\begin{center}
\begin{tabular}{c c c}
\hline
Class & Acc.(\%) & F1(\%) \\
\hline
Angry & 77.07 & 76.02 \\
Happy & \underline{55.95} & \underline{65.52}  \\
Neutral & 82.10 & 79.29 \\
Sad & 80.00 & 77.54 \\
\hline
\end{tabular}
\label{tab:each}
\end{center}
\end{table}

\begin{figure}
    \centering
    \includegraphics[width=0.8\linewidth]{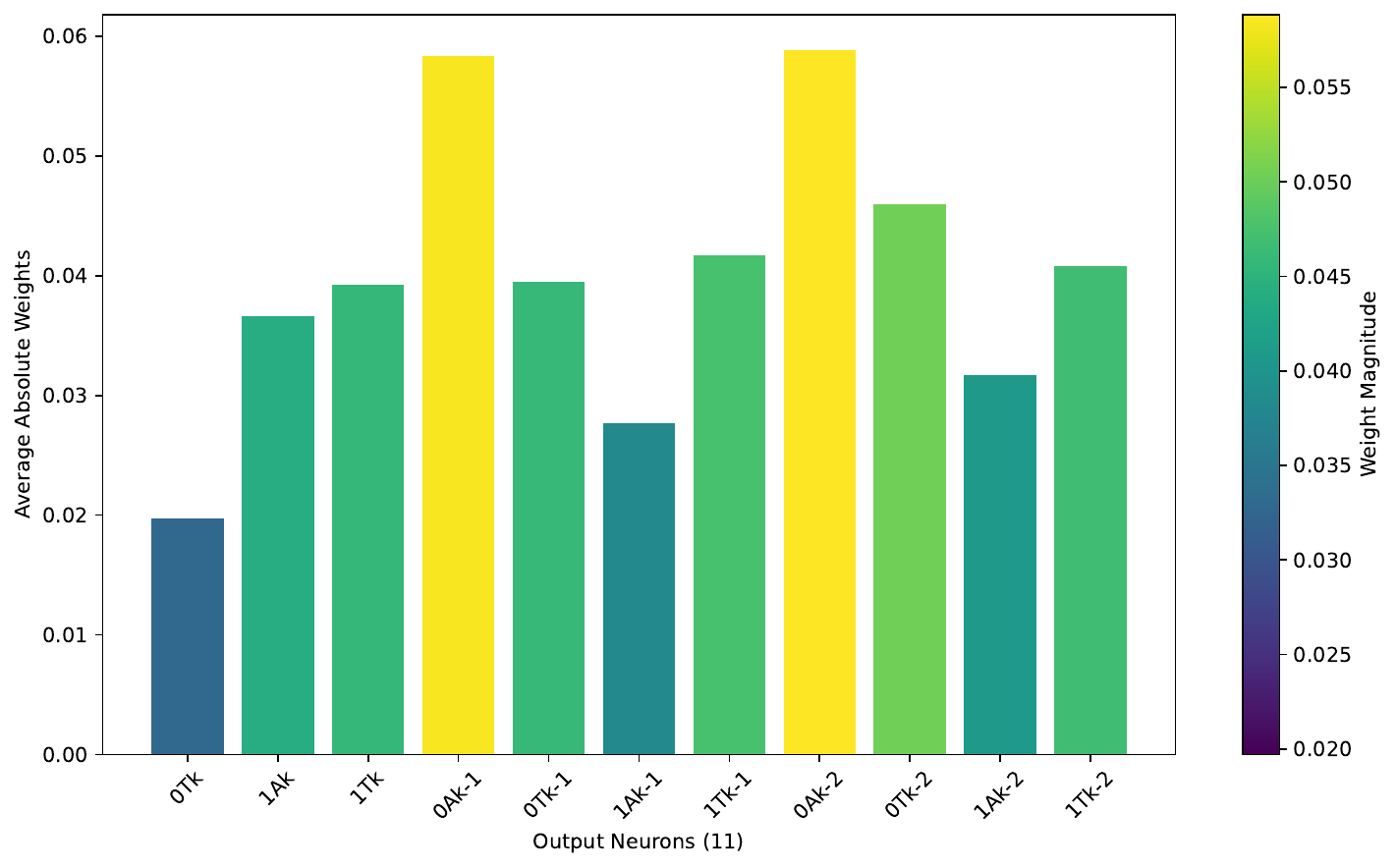}
    \caption{Average absolute weights for each output neuron. `0' represents the current speaker, `1' represents the interlocutor. `A' denotes audio, `T' denotes text, and `k' denotes the time frame.}
    \label{fig:weight}
    \vspace{-0.2cm}
\end{figure}

Additionally, one of the key contributions of our model is its ability to explain the importance of each input feature. Using the 48-seed model, the weights are visualised in Figure \ref{fig:weight}. This visualisation provides interpretability, demonstrating that audio has a greater influence on the ERC tasks, which aligns with our intuition. We think it is due to the presence of more acoustic features, such as pitch and tone variations, 
whereas text is limited to key words. It gives us also the insight that xLSTM may be better at processing audio features. Furthermore, the analysis reveals that the audio from the interlocutor has a relatively smaller influence compared to the speaker's own audio. However, the interlocutor plays an important role in emotional contagion and affective interaction 
\cite{hatfield1993emotional}, which is the reason we add them to our feature sets.

We conducted an ablation study to systematically evaluate the contributions of each component in our framework toward achieving SOTA performance. Our framework incorporates three key mechanisms: CLAP for multimodal feature alignment, a Gated mechanism for weighted feature integration, and DED as a post-adjustment module. The results of this ablation study are presented in Table \ref{tab:ablation study}.

The experimental results demonstrate that the CLAP feature alignment mechanism improves performance over an IAAN model, underscoring the importance of effectively aligning multimodal features. Further gains are achieved by incorporating the Gated-xLSTM architecture, which enhances the model's ability to selectively integrate features. Compared to the base xLSTM, the Gated-xLSTM with CLAP achieves a substantial improvement, with weighted accuracy increasing by 6.64\% and weighted F1-score by 7.55\%. This indicates that the gating mechanism effectively refines feature representations. Moreover, performance improves with the addition of the DED post-adjustment module, which fine-tunes the model's predictions. This final configuration achieves the highest performance, with weighted accuracy and weighted F1-score increasing by 9.08\% and 9.88\%, respectively. The incremental improvements at each stage of the ablation study validate the effectiveness of our design choices and highlight the synergistic benefits of combining these mechanisms.

\begin{table}[htbp]
     \caption{Ablation study for each components we utilise: CLAP, Gate mechanism, and DED}
     
    \centering
    \resizebox{0.45\textwidth}{!}{
    \begin{tabular}{c c c c c}
    \hline
    Feature & Model & Post Adjustment & W-Acc & W-F1 \\
    \hline

    IAAN         & base xLSTM         & -         & 67.26           & 66.09        \\ 
CLAP         & base xLSTM         & -         & 69.93$_{({\uparrow}2.67)}$ & 69.76$_{({\uparrow}3.67)}$ \\ 
CLAP         & Gated-xLSTM        & -         & 73.90$_{({\uparrow}6.64)}$ & 73.64$_{({\uparrow}7.55)}$ \\ 
CLAP         & Gated-xLSTM        & DED       & \textbf{76.34}$_{({\uparrow}9.08)}$ & \textbf{75.97}$_{({\uparrow}9.88)}$ \\ 
    \hline
    \end{tabular}
}
    \label{tab:ablation study}
    \vspace{-0.3cm}
\end{table}


\section{Conclusion}
In this study, we proposed a novel framework based on GatedxLSTM, which is the first to employ xLSTM-based models in the field of MER. We have designed a pipeline that incorporates multimodality alignment, selective and concentrated feature extraction, and a post-adjustment mechanism. This framework achieves SOTA performance in four-class emotion recognition on the popular IEMOCAP dataset. Furthermore, our project enhances the interpretability of model decisions from a psychological perspective, providing deeper insights into the mechanisms of affective computing. These insights offer potential inspiration for the future development of AC. In future work, we aim to expand our model research to real-time emotion recognition, with the goal of advancing AC intelligence and fostering the development of more user-friendly AGI.

\section*{Ethic Impact Statement}
Our research advances affective computing by improving the accuracy and explainability of emotion recognition systems. By leveraging state-of-the-art methodologies, our work enhances model interpretability, ensuring that predictions are more transparent and trustworthy. This contributes to the development of fairer and more reliable emotion recognition technologies, supporting applications in healthcare, human-computer interaction, and assistive technologies. Our approach does not introduce any obvious ethical concerns.
\bibliographystyle{IEEEtran}
\bibliography{ref}
\end{document}